\documentclass[11pt,a4paper]{article}
\usepackage{hyperref}
\usepackage[hyperref]{acl2020}
\usepackage{times}
\usepackage{latexsym}

\usepackage{microtype}

\aclfinalcopy 

\usepackage{verbatim}
\usepackage{xspace}
\usepackage{booktabs}
\usepackage{graphicx}
\usepackage{amssymb}
\usepackage{multirow}
\usepackage{stackengine}

\usepackage{adjustbox}
\usepackage{array}

\newcolumntype{R}[2]{%
    >{\adjustbox{angle=#1,lap=\width-(#2)}\bgroup}%
    l%
    <{\egroup}%
}
\newcommand*\rot{\multicolumn{1}{R{45}{1em}}}

\usepackage{url}

\title{The Dialogue Dodecathlon: \\
Open-Domain Knowledge and Image Grounded Conversational Agents}

\author{
  \qquad \quad Kurt Shuster, Da Ju, Stephen Roller \\
  \qquad \quad \textbf{Emily Dinan, Y-Lan Boureau, Jason Weston} \\
  \qquad \quad Facebook AI Research \\
  \qquad \quad {\tt \{kshuster,daju,roller,edinan,ylan,jase\}@fb.com}
}
\date{}

\begin{document}
\maketitle
\begin{abstract}
We introduce 
{\em dodeca}Dialogue:  a set of 12 tasks that measures
if a conversational agent can communicate engagingly with personality and empathy, ask questions, answer questions by utilizing knowledge resources, discuss topics and situations,
and perceive and converse about images. By multi-tasking on such a broad
large-scale set of data, we hope to both move towards and measure progress in producing
a single unified agent that can perceive, reason and converse with humans in an open-domain setting. We show that such multi-tasking improves over a BERT pre-trained baseline, largely 
due to multi-tasking with very large dialogue datasets in a similar domain, and that the multi-tasking in general provides gains to both text and image-based tasks using several metrics in both the fine-tune and task transfer settings. We obtain state-of-the-art results on many of the tasks, providing a strong baseline for this challenge. 
\end{abstract}

\section{Introduction}

One of the goals of AI is to build a seeing, talking agent that can discuss, reason, 
empathize, and provide advice  -- in short a system that can perform natural communication 
displaying many of the properties expected when speaking to a human partner.
Ideally, it should be able to be knowledgeable and personable, expert and engaging,
serious or humorous -- depending on the situation. It should be capable of answering questions,
asking questions, responding to statements, having its own persona, and grounding the dialogue with external information and images. 

While  no single  task exists that can train an agent or measure its ability 
on all of these axes at once, a number of distinct large-scale datasets targeting subsets of these skills have recently become available.
We thus assemble these disparate tasks to form a single challenge: {\em dodeca}Dialogue, consisting of 12 subtasks. Each contains both training data to build the skills we desire for our agent, and validation and test sets to measure our agent's ability at that skill. The overall goal is a single agent that can display all these skills.
As some of the subtasks have very large datasets, e.g. 2.2 billion utterances, they can possibly help the agent with other skills too.

We thus build a model capable of training and multi-tasking on all these sources. We employ a transformer-based architecture \citep{vaswani2017attention} which accepts an image, external textual information and dialogue history as input, and generates a response for a given dialogue turn.
Practically, by pre-training on the largest of the subtasks and then multi-tasking on all them, we can obtain state-of-the-art results compared to existing independently reported  performance on all 10 of the 12 subtasks that have previous comparable results. We hence set a strong baseline for this challenge. While many existing approaches use large-scale pre-training on general text corpora, we show that using dialogue datasets instead, which are more closely linked to the desired agent's goals, is a strong alternative.

However, many challenges remain. 
While multi-tasking performs well, and has clear benefits,
as shown in other works  \citep{liu2015representation,raffel2019exploring}, when compared to fine-tuning of the same system we do obtain typically small losses. 
Zero-shot transfer to left-out tasks 
is also demanding for current approaches.
We analyze these aspects, along with our model's ability to ground on external knowledge and images in conjunction with the dialogue context, the impact of decoding algorithms, analysis of the weighting of tasks during multi-tasking as well as cross-task transfer ability in order  to shed light and make progress on this challenging topic.

\begin{table*}
    \centering
   \begin{small}
    \begin{tabular}{llllllllrrrrr}
&
\rot{\multirow{2}{*}{Ask Questions}} & 
\rot{\multirow{2}{*}{Answer Questions}} &                                                    
\rot{\multirow{2}{*}{Respond to Statements}} & 
\rot{\multirow{2}{*}{Persona Grounding}} & 
\rot{\multirow{2}{*}{Knowledge Grounding}} & 
\rot{\multirow{2}{*}{Situation Grounding}} & 
\rot{\multirow{2}{*}{Image Grounding}} & 
\multicolumn{4}{l}{} & Resp.\\
Name   &
& & & & & & &
Train & Valid & Test &                                                      
{\# Turns} & 
{Length}
\\
\toprule
 ConvAI2           & \checkmark & \checkmark & \checkmark & \checkmark &            &            &            & 131,438       & 7,801  & 6,634  & 14.8   & 11.9     \\
 DailyDialog       & \checkmark & \checkmark & \checkmark &            &            &            &            & 87,170        & 8,069  & 7,740  & 7.9 & 14.6  \\
 Wiz. of Wikipedia & \checkmark & \checkmark & \checkmark &            & \checkmark &            &            & 74,092        & 3,939  & 3,865  & 9.0   & 21.6     \\
 Empathetic Dialog & \checkmark & \checkmark & \checkmark &            &            & \checkmark &            & 40,252        & 5,736  & 5,257  & 4.3 & 15.2  \\
 Cornell Movie     & \checkmark & \checkmark & \checkmark &            &            &            &            & 309,987       & 38,974 & 38,636 & 4.0   & 15.0   \\
 LIGHT             & \checkmark & \checkmark & \checkmark & \checkmark &            & \checkmark &            & 110,877       & 6,623  & 13,272 & 13.0   & 18.3     \\
 ELI5              &            & \checkmark &            &            & \checkmark &            &            & 231,410       & 9,828  & 24,560 & 2.0   & 130.6 \\
 Ubuntu            & \checkmark & \checkmark & \checkmark &            &            &            &            & 1,000,000     & 19,560 & 18,920 & 2.0   & 18.9     \\
 Twitter           & \checkmark & \checkmark & \checkmark &            &            &            &            & 2,580,428     & 10,405 & 10,405 & 2.0   & 15.7     \\
 pushshift.io Reddit            & \checkmark & \checkmark & \checkmark &            &            &            &            & $\sim$ 2200 M & 10,000 & 10,000 & 2.0   & 35.0     \\
 Image Chat        & \checkmark & \checkmark & \checkmark & \checkmark &            &            & \checkmark & 355,862       & 15,000 & 29,991 & 3.0   & 11.4     \\
 IGC               & \checkmark & \checkmark &            &            &            &            & \checkmark & 4,353         & 486    & 7,773  & 3.0   & 8.6     \\

\bottomrule
    \end{tabular}  
    \end{small}
    \caption{The 12 {\em dodeca}Dialogue subtasks, their sizes (number of train, valid, test utterances), and average number of turns and response length (words).}
    \label{tab:datasets}                                                    
\end{table*}

\section{The {\em dodeca}Dialogue Task}

The {\em dodeca}Dialogue task is intended to assemble important aspects of an engaging
conversational agent into a single collection, 
where each subtask covers some of those goals.
Such an  agent should be able to get to know you when you first talk to it (ConvAI2), discuss everyday topics (DailyDialog, pushshift.io Reddit, Twitter, Cornell Movie), speak knowledgeably at depth (Wizard of Wikipedia, Ubuntu) and answer questions on such topics (ELI5). It must be able to handle situated conversations and demonstrate empathy (Empathetic Dialog, LIGHT) . It can also discuss images, as this is a vital part of human connection (Image Chat, IGC). We note that all of the provided subtasks are in English.

The overall statistics of the subtasks are given in Table \ref{tab:datasets}.
We now discuss each in turn.

\paragraph{ConvAI2}

ConvAI2 is a dataset used at the NeurIPS 2018 competition of the same name, and is based on PersonaChat \citep{zhang2018personalizing, dinan2019second}.
The training data involves
paired crowdworkers having a conversation where they get to know each other, in which each is given a role to play based on sentences describing their persona, which were also separately crowdsourced (while they cannot see their partner's persona).  
It thus involves asking and answering questions, responding in kind, and getting
to know the other speaker and engaging 
them in friendly 
conversation -- useful skills for an open-domain conversational agent.

\paragraph{DailyDialog}

\citet{dailydialogue} built a dialogue dataset intended to reflect conversations occurring in daily  life. 
It covers  ten categories ranging from holidays to financial topics, rather than focusing on one domain. Compared to ConvAI2, these conversations seem more in keeping with partners who already know each other, and want to discuss typical life details, again useful skills for a conversational agent.
The dataset is also annotated with topic, emotion and utterance acts, but here we ignore these annotations and learn only from the utterances in the dialogue turns.

\paragraph{Wizard of Wikipedia}
This task involves discussing a given topic in depth, 
where the goal is to both engage the partner 
as well as display expert knowledge \citep{dinan2018wizard}.
The training set consists of 1247 topics and a retrieval system over Wikipedia from which the dialogues were grounded during the human-human crowdsourced conversations. 
The topics were also crowdsourced and range from e-books to toga parties to showers.
A model can thus learn to also perform similar retrieval and grounding at test time to potentially discuss any topic if it can generalize. 
We use the gold knowledge version of the task. 
We see this skill as a core component of an agent being able to not just chitchat, but 
actually engage a user in discussing real information about the world, e.g. by retrieving over
documents from the internet.

\paragraph{Empathetic Dialogues}
\citet{rashkin2019empathy} constructed a dataset of
crowdworker conversations grounded in 
an emotional situation. In each dialogue, one speaker describes a personal situation and the other plays a ``listener'' role, displaying empathy during the discussion. The dataset contains descriptions of the situations being discussed with an attached emotion label, but these are not used here.
Trained models are measured playing the part of the empathetic listener, an important
feature of an agent to which humans wish to speak. 

\paragraph{Cornell Movie}
\citet{cornell} constructed a  corpus containing a collection of fictional conversations from movie scripts, thus covering a large diversity of topics and emotional states.

\paragraph{LIGHT}

LIGHT \citep{urbanek2019learning} involves situated interactions between characters in a text adventure game. Similar to ConvAI2, personas for each character are given, with the training set including conversations between crowdworkers playing those roles.
Different from ConvAI2, included are emotes and actions
grounded within the game world (e.g. picking up and giving objects).  
As such, it measures the ability of a conversational agent to ground its
discussion on a dynamic environment.

\paragraph{ELI5}
ELI5 \citep{fan2019eli5} involves long-form question answering grounded on multiple retrieved documents in order to answer common questions which people ask on the popular ELI5 subreddit.
As such, the answers are in a conversational form applicable to a dialogue agent.

\paragraph{Ubuntu}
\citet{ubuntu} built a dataset that involves in-depth discussions in solving Ubuntu problems. This studies the ability of an agent on a very focused single topic, and is also a standard benchmark in the field.               

\paragraph{Twitter}
We use a variant of Twitter discussions (text-only),  which have been used in many existing studies, e.g. \citet{twitter,see2019makes}. This data naturally involves everyday discussions about topics that people care about. The public forum makes them different from the more personal discussions of some of the other tasks. This is the second largest dataset in the collection, and we thus measure in experiments its ability to help performance on other tasks.

\paragraph{pushshift.io Reddit}
We use a variant of Reddit discussions (text-only), which has also been used in several existing studies, see e.g. \citet{reddit_use, mazare2018training,keskar2019ctrl}.
Following \citet{humeau2019real}, we use a previously existing Reddit dataset extracted and obtained by a third party and made available on pushshift.io, training to generate a comment conditioned on the full thread leading up to the comment, spanning 2200M training examples.
This is the largest dataset in the collection -- much larger than the others. The subreddits cover a vast range of topics, and hence this is a strong candidate for helping
improve performance on other tasks via pre-training and multi-tasking.  
Note this dataset does not overlap with ELI5.

\paragraph{Image Chat}
\citet{shuster2018imagechat} collected a crowdsourced dataset of human-human conversations
about an image with a given personality, where the goal is to engage the other 
speaker. As such, it covers natural conversational responses, including displays of emotion and humor.

\paragraph{Image Grounded Conversations (IGC)}
IGC \citep{igc} similarly involves two speakers discussing an image, here focusing on
questions and responses. It only includes a validation and test set, and so we converted
most of the validation set to form a small training set.


\subsection{Evaluation}

\paragraph{Metrics}
For all tasks, we use the following metrics: perplexity (PPL), BLEU,  
ROUGE-1,-2 and -L and F1, and also pick the metric most used in the literature
as that subtask's `Score' to compare to existing work.

\paragraph{Multi-tasking}
As we are interested in building a single conversational agent, we measure the ability of multi-tasked models that can perform all twelve tasks at once.

\paragraph{Single-Task Fine-tuning}
We can still compare such multi-tasked models to single-task fine-tuned baselines to assess if we have gained or lost performance. Like other works \citep{liu2015representation,raffel2019exploring} we also consider a multi-task followed by finetune setup in order to see if this produces better models. The latter tests if multi-tasking still proves useful in the single-task setting.

\paragraph{Zero-shot Transfer}
Finally, we consider a leave-one-out zero-shot setting whereby training is constrained to be on all the training data {\em except for the task being evaluated}. This evaluates the performance
on truly new unseen tasks, an important behavior given there are always new tasks.

\section{Related Work}

\subsection{Existing Models and Results}

Where possible, we have tried to track the best existing results for each task
and provided a comparison in our final results table.

As ConvAI2 was a competition, a number of competitors built strong models on it.
The best results were obtained by large pre-trained transformers \citep{dinan2019second}. In particular, \citet{wolf2019transfertransfo} pre-trained
via the method of \citet{radford2018improving} using the BooksCorpus dataset, 
resulting in the best perplexities and F1 scores.
Since then, results have gotten even better with the advent of better and larger pretraining
\citep{lewis2019bart}, which we compare to here; the same work also reports strong results on ELI5.

\citet{mixreview} recently obtained strong results on the DailyDialog and Cornell Movie tasks in terms of perplexity by pre-training on 10\% of CC-NEWS \citep{bakhtin2019real}, thus using 100 million sentences (2.7 billion words) and then fine-tuning a transformer based model with a multi-task strategy.

Overall, large pre-trained transformers indeed provide strong existing results on many
of the tasks.
Several large language modeling projects have been undertaken in order to
show prowess in multi-tasking ability \citep{radford2019language,keskar2019ctrl},
and transformer-based approaches have been adapted to language and vision tasks as well \cite{lu2019vilbert,tan2019lxmert,li2019visualbert,shuster2018imagechat}. 
As well as citing the relevant papers' results where possible in the experiments section, we also
train a BERT-based \cite{devlin2018bert} generative model as an additional baseline.

\subsection{Related Tasks and Collections}

In the interests of feasibility, there are tasks we did not include in 
{\em dodeca}Dialogue. For example, there are additional
knowledge tasks \citep{qin2019conversing,moghe2018towards,Gopalakrishnan2019}
and image-based datasets \citep{das2017visual} one could use. There are also a large number of QA tasks we did not include, e.g. \citet{rajpurkar2016squad,choi2018quac}. In general, our choices were made based on tasks that after training 
might produce an engaging dialogue agent that humans naturally would want to talk to -- which means either natural datasets or crowdsourced datasets where crowdworkers were encouraged to
engage one another.  As computational resources and 
ambitions scale, it would be interesting to add more tasks as well, while retaining
the twelve we have chosen here in order to continue to evaluate their success, whilst 
extending the scope of the entire system.

All the subtasks in the collection we use here already exist.
Other research projects have also built such collection-based tasks before as well.
In particular, the NLP decathlon \citep{mccann2018natural}, 
from which the name of this paper is inspired,
collects together a diverse set of NLP tasks -- from sentiment detection to parsing.
\citet{talmor2019multiqa} collect a set of 10 QA datasets and build {\sc MultiQA}.
Recently, \citep{raffel2019exploring} also similarly multi-tasked a large set of NLP tasks, on an even bigger scale. Our work differs from these 
in that it is focused on dialogue tasks which naturally group together to form a
conversational agent.

\setlength{\tabcolsep}{9pt}
\begin{table*}
    \centering
    \begin{small}
    \begin{tabular}{l|r|rrrrrr|rr}
\rot{}   &
\rot{BERT-based} & 
\rot{Single Task (from scratch)} & 
\rot{Single Task (fastText init)} & 
\rot{Twitter + Single Task} & 
\rot{Reddit Only} & 
\rot{Reddit + Single Task} & 
\rot{MT All Tasks + FT Single Task} & 
\rot{All Tasks MT} & 
\rot{\stackanchor{Leave-One-Out}{Zero-Shot}} \\
\toprule
ConvAI2 & 19.4 & 43.3 & 38.9 & 28.7 & 18.3 & 11.4 & \textbf{11.2} & 11.3 & 16.4\\ 
DailyDialog & 15.2 & 37.8 & 32.8 & 20.8 & 18.2 & 10.4 & \textbf{10.2} &  11.8 & 15.5\\ 
Wiz. of Wikipedia & 14.1 & 40.7 & 36.0 & 37.3 & 15.3 & 8.7 & \textbf{8.5} &  8.7 & 13.2\\ 
Empathetic Dialog & 23.2 & 47.1 & 40.5 & 23.1 & 14.4 & 11.3 & \textbf{11.1} & 11.2 & 13.0\\ 
Cornell Movie & 29.4 & 46.2 & 44.8 & 34.2 & 27.8 & 20.0 & \textbf{19.8} & 22.3 & 25.4\\ 
LIGHT & 29.7 & 63.6 & 57.5 & 40.0 & 32.9 & \textbf{18.7} & \textbf{18.7} & 19.0 & 26.9\\ 
ELI5 & 28.1 & 62.9 & 58.8 & 63.8 & 31.2 & 21.2 & \textbf{21.1} & 25.0 & 31.1\\ 
Ubuntu & 20.7 & 35.8 & 34.5 & 38.5 & 31.1 & 17.3 & \textbf{17.2} &  23.3 & 30.8\\ 
Twitter & 37.0 & 61.9 & 59.3 & 59.3 & 53.6 & \textbf{29.8} & \textbf{29.8} & 37.0 & 52.8\\ 
pushshift.io Reddit & 39.0 & 27.8 & 27.8 & 27.8 & 27.8 & 27.8 & \textbf{25.8} & 28.0 & 106.3\\ 
Image Chat & \tiny{N/A} & 40.1 & 37.4 & 31.1 & 32.5 & \textbf{18.3} &\textbf{ 18.3} & 21.8 & 29.3\\ 
IGC & \tiny{N/A} & 86.3 & 79.5 & 23.1 & 14.6 & \textbf{10.0 }& \textbf{10.0} &  10.2 & 12.2\\ 
\midrule
{\em dodeca}Score & \tiny{N/A} & 49.5 & 45.7 & 35.6 & 26.5 & 17.1 & \textbf{16.8} & 19.1 & 31.1\\
\bottomrule
    \end{tabular}
    \end{small}
    \caption{Validation perplexity for the {\em dodeca}Dialogue tasks in various settings.}
    \label{tab:valid_all}
\end{table*}

\section{Models}

\paragraph{BERT baseline.} We implement a generative baseline using BERT via adapting the model using a standard auto-regressive loss. We concatenate both the context and current generation and provide these as input to the model, using BERT's sentence embeddings to distinguish the roles in the network. Although BERT is trained to predict masked tokens, we find that fine-tuning can easily adjust its behavior to predicting the next token. Our BERT baseline is roughly equivalent to the model of \citet{wolf2019transfertransfo}, but does not have a classification loss term. The implementation relies on HuggingFace Transformers \cite{wolf2019huggingfaces}. 
We thus fine-tune this model for each of our tasks, except Image Chat and IGC which require images as input.

\paragraph{Image+Seq2Seq.} 
We use a modification of a transformer Seq2Seq architecture 
\cite{vaswani2017attention}, additionally adding image features to the encoder.
 Our model is a 8 layer encoder, 8 layer decoder with 512 dimensional embeddings 
 and 16 attention heads, and is based on the ParlAI  implementation  \citep{miller2017parlai}. We use BPE following \citet{humeau2019real} via 
  lower-cased Wikipedia, Toronto Books, and Open
Subtitles with 30k merges, giving 54,940 terms. Reported perplexities are computed with this dictionary.
 For image features, we use the pre-trained image features from the ResNeXt-IG-3.5B model, a ResNeXt 32 x 48d architecture \citep{resnextXie} trained on 3.5 billion Instagram images following the procedure described by \citet{uru}. This model was previously used successfully for the Image Chat task in \citet{shuster2018imagechat}. The final encoding from the ResNeXt   model is a vector of size 2048; we then use a linear layer to project into the same size as the text encoding, and add it as an extra token at the end of the transformer's encoder output, then feed them all into the decoder. During fine-tuning we train the text transformer, but leave the image encoding fixed, apart from fine-tuning the linear projection. 
 The text transformer is fine-tuned with a standard auto-regressive negative log-likelihood (NLL) loss, following usual sequence to sequence training schemes.
 
Our best models are available at \url{parl.ai/projects/dodecadialogue}.

\section{Experiments}
\label{sec:experiments}

\paragraph{Task Training}
We employ the ParlAI framework \citep{miller2017parlai} for training on single tasks and for 
multi-tasking, as many of the tasks are already implemented there, along with a (multi-task) training and evaluation framework for such models.

\paragraph{Pre-training} As pushshift.io Reddit and (to some extent) Twitter are much larger than our other tasks,
we try pre-training the Seq2Seq module of our Image+Seq2Seq networks with those datasets, before multi-tasking on all of the tasks, or for evaluating single task fine-tuning.

For Reddit, the model was trained to generate a comment conditioned on the full thread leading up to the comment. Comments containing URLs or that were under 5 characters in length were removed from the corpus, as were all child comments. Comments were truncated to 1024 BPE tokens.
The model was trained with a batch size of 3072 sequences for approximately 3M updates using a learning rate of 5e-4, and an inverse square root scheduler. This took approximately two weeks using 64 NVIDIA V100s.
We note that our transformer pre-training only includes text, while our 
image encoder was pre-trained separately in previous work \citep{uru}.
Learning how to combine these sources occurs during fine-tuning.

It is important to note that, while compute-heavy, pre-training was conducted exactly once, and all of the subsequent fine-tuning is significantly faster to run.

\begin{table}
    \resizebox{0.95\linewidth}{!}{
    \begin{tabular}{lrrr}
Model &    \rot{ConvAI2} & \rot{Wiz. of Wikipedia} & \rot{\stackanchor{Empathetic}{Dialog}} \\ 
\hline
Reddit            & 18.3 &  15.3 & 14.4 \\ 
Reddit+ConvAI2           & {\textbf{11.4}} &  14.2 & 14.7 \\
Reddit+Wiz. of Wikipedia & 16.3 &  \textbf{{8.7}}  & 14.0 \\
Reddit+Empathetic Dialog & 17.9 &  15.3 & 11.3 \\
Multi-Tasking All 4 Tasks& 11.6 &  \textbf{{8.7}}  & \textbf{{11.2}} \\
    \end{tabular}
    }
    \caption{Transfer performance of various multi-task models (validation perplexity).   }
    \label{tab:transfer_some}
\end{table}

\paragraph{Transfer Performance between Tasks} 

We first perform a preliminary study on a subset of the tasks:
Reddit, ConvAI2, Wizard of Wikipedia and Empathetic Dialogues,
and report the transfer ability of training on some of them, 
and testing on all of them (using the validation set), reporting 
perplexity.
The results are reported in Table \ref{tab:transfer_some}.
They show that training on pushshift.io Reddit alone, a huge dataset, is effective at transfer to
other tasks, but never as effective as fine-tuning on the task itself.
Moreover, fine-tuning on most of the smaller tasks actually provides improvements
over pushshift.io Reddit training alone at transfer, likely because the three tasks selected are more
similar to each other than to pushshift.io Reddit. Finally, training on all four tasks
is the most effective strategy averaged over all tasks compared to any other single model,
although this does not beat switching between different fine-tuned models on
a per-task basis.

\setlength{\tabcolsep}{8pt}
\begin{table}[t]
    \centering
    \begin{small}
    \begin{tabular}{lrr}
{\em Knowledge grounding} & Without & With\\
Wiz. of Wikipedia   & 16.8   & \textbf{8.7} \\
ELI5              & 21.3   & \textbf{21.2}\\ 
\hline
{\em Image grounding}\\
Image Chat           & 19.5 & \textbf{18.3} \\ 
IGC                 & \textbf{10.1} & \textbf{10.1} \\ 
    \end{tabular}
    
    \end{small}
    \caption{The impact of knowledge and image grounding in {\em dodeca}Dialogue (validation perplexity). }
    \label{tab:grounding}
\end{table}

\paragraph{Comparison of Pre-training + Fine-tuning strategies} 
Across all 12 tasks, we compare several pre-training strategies:  using BERT, no pre-training at all, only initializing via fastText \citep{fasttext}, 
 and using Twitter and pushshift.io Reddit pre-training with our Image+Seq2Seq architecture.
For each variant we tune the learning rate, layers, number of heads and embedding size,
with less pre-training typically requiring smaller capacity models. 
We then only fine-tune on a single task in these experiments, and report perplexity for that task alone, over all 12 tasks.
The results are given in Table \ref{tab:valid_all}, reporting results on the validation 
set\footnote{We choose not to use the test set here as we report so many numbers, we do not want to overuse it.}.

The results show a clear reduction in perplexity with more pre-training, as expected. This is most easily seen by the {\em dodeca}Score (last row) that is the mean perplexity over all 12 tasks, which decreases from 49.5 (from scratch models) down to 17.1 with pushshift.io Reddit pre-training.
FastText (45.7) and Twitter (35.6) initializations help, but nowhere near as much.
BERT fares better, but still is clearly worse than pushshift.io Reddit pre-training. The hypothesis here is that pushshift.io Reddit yields much more effective transfer as it is a dialogue task like our others, whereas non-dialogue corpora such as Wikipedia are not. This was previously observed for retrieval models in  \citet{humeau2019real}. Note that we do not report results for the image dialogue tasks for BERT as that architecture does not deal with images.

Finally, as pushshift.io Reddit is so effective, we also compare to pushshift.io Reddit training only, with no fine-tuning at all across all tasks, similar to our initial study in
Table \ref{tab:transfer_some}. The performance is impressive, with some tasks yielding lower perplexity than BERT pre-training + single task fine-tuning. However, it still lags 
significantly behind fine-tuning applied after pushshift.io Reddit pre-training.


\setlength{\tabcolsep}{5.5pt}
\begin{table}[t]
    \centering
    \begin{small}
      \begin{tabular}{l|rrrrrrr} 
   &  \multicolumn{6}{c}{Relative Task Weighting} \\
    &  1  & 2 & 5 & 10 & 20 & 50 & $\infty$\\
Cornell  & 21.9 & 21.5 & 20.6 & 20.1 & 19.9 & \textbf{19.8} & -\\
{\em\tiny Fine-tuned} & 20.1 & 20.0 & 20.0 & 19.9 & \textbf{19.8} & \textbf{19.8} & 20.0\\
\hline
ELI5 & 25.0 & 24.1 & 22.8 & 22.2 & 21.6 & 21.3 & -\\
{\em\tiny  Fine-tuned}  & 21.8 & 21.6 & 21.4 & 21.3 & \textbf{21.1} & \textbf{21.1} & 21.2\\
\hline
Ubuntu  & 23.1 & 22.2 & 20.6 & 19.6 & 18.6 & 17.4 & - \\ 
{\em\tiny Fine-tuned} & 18.2 & 18.1 & 17.8 & 17.7 & \textbf{17.2} & \textbf{17.2} & 17.3\\
    \end{tabular}
    \end{small}
    \caption{Validation perplexity on select {\em dodeca}Dialogue tasks comparing relative weights of tasks during multi-tasking, followed by fine-tuning (row below). The relative task weight is the ratio of examples from that task compared to others presented during multitasking. $\infty$ indicates single-task training.} 
    \label{tab:mt_ft_weighting_res}
\end{table}

\setlength{\tabcolsep}{5pt}
\begin{table}
    \centering
    \begin{small}
    \begin{tabular}{lrrrrrc}
                    & & & & &N-gram &  \\ 
                    &  \multicolumn{4}{c}{Beam Size} & \multicolumn{1}{c}{Block} &  Nucleus \\
Task                &  1 &    2    & 3    & 5    &  $N=3$ & $p=$0.3  \\
\hline
ConvAI2             &  20.0 & 21.0 & \textbf{21.3} & 21.2 &  \textbf{21.3} & 18.7   \\
WoW                 &  35.9 & 37.4 & 37.8 & \textbf{37.9} & \textbf{ 37.9} & 31.1   \\
    \end{tabular}
    \end{small}
    \caption{Impact of the decoding strategy on select tasks, reporting validation F1 score for the All Tasks MT model. N-gram block is for best beam size.  }
    \label{tab:decoding_res}
\end{table}

\paragraph{Image and Knowledge Grounding} 
 
Some of our tasks involve grounding on knowledge or images.
To show such grounding helps, we report results with and without grounding on
those tasks in Table \ref{tab:grounding}, reporting perplexity. 
Particularly for Wizard of Wikipedia
(knowledge) and Image Chat (images) such grounding has a clear effect.

\setlength{\tabcolsep}{8pt}
\begin{table*}[t]
    \centering
    \begin{small}
    \begin{tabular}{l|rrrl||rr||rr}
   & \multicolumn{4}{c}{Existing Approaches (independent)} &  \multicolumn{2}{c}{MT + FT  } & \multicolumn{2}{c}{All Tasks MT } \\
   & Approach & PPL & Score & \small{(Metric)} &  PPL  & Score &  PPL  & Score\\
ConvAI2 & \tiny{\cite{lewis2019bart}} & *11.9 & *20.7 & \small{F1} & 11.1 & 21.6 & \textbf{10.8} & \textbf{21.7} \\
DailyDialog & \tiny{\citep{mixreview}} & 11.1 & - & \small{F1} & \textbf{10.4} & \textbf{18.2} &  12.0  &  16.2  \\
Wiz. of Wikipedia & \tiny{\citep{dinan2018wizard}}  & 23.1  & 35.5 & \small{F1} & \textbf{8.3} & \textbf{38.4} &  8.4  &  \textbf{38.4}  \\
Empathetic Dialog & \tiny{\citep{rashkin2019empathy}} & 21.2 & 6.27 & \tiny{Avg-BLEU} & \textbf{11.4} & 8.1 & 11.5 & \textbf{8.4} \\
Cornell Movie & \tiny{\citep{mixreview}} & 27.5 & - & \small{F1} & \textbf{20.2} & \textbf{12.4} &  22.2  &  11.9   \\
LIGHT & \tiny{\cite{urbanek2019learning}} & $^*$27.1 &  $^*$13.9 & \small{F1} & \textbf{18.9} & \textbf{16.2} & 19.3  & 16.1  \\
ELI5 & \tiny{\cite{lewis2019bart}} & 24.2  & 20.4 & \tiny{Avg-ROUGE} & \textbf{21.0} & \textbf{22.6} & 24.9 & 20.7 \\
Ubuntu & \tiny{\citep{DBLP:journals/corr/LuanJO16}} & 46.8 & - & \small{F1}  & \textbf{17.1} & 12.7 &  23.1  &  12.1  \\
Twitter & & -  & -  & \small{F1} & 30.7 & 9.9 &  38.2  & 9.8  \\
pushshift.io Reddit & & - & - & \small{F1} & 25.6 & 13.6 &  27.8  &  13.5  \\
Image Chat  & \tiny{\cite{shuster2018imagechat}} & - & 27.4  & \tiny{ROUGE-L (1$^{st}$ turn)} & \textbf{18.8} & \textbf{43.8} &  22.3  &  39.7  \\
IGC & \tiny{\citep{igc}} & - & 1.57 & \tiny{BLEU (responses)} & 11.9 & \textbf{9.9} & 12.0  & 8.2 \\
\hline
    \end{tabular}
    \end{small}
    \caption{Test performance for various metrics on the {\em dodeca}Dialogue tasks comparing our multi-task and multi-task + fine-tuned methods to existing approaches (cited). Dashes mean metric was not provided. $^*$ was reported on validation only. Score is defined on a per-task basis in the metric column.} 
    \label{tab:sota}
\end{table*}

\paragraph{Multi-Task Results} 

Next, we perform multi-task training across all tasks, which 
is our ultimate goal in order to obtain an open-domain conversational agent.
We optimize over the same set of hyperparameters as before, including 
multi-tasking weights for tasks, where one samples during training with differing probabilities, and we choose the best model by performing early stopping on the average performance across all tasks. In this way, we treat all 12 tasks as a single task, and thus during test time it is the model's responsibility to understand how to respond from the context (image/dialogue) itself.

In the end we did not obtain
clear improvements beyond pre-training with pushshift.io Reddit and then equally sampling from all tasks.
We report that final model's validation performance in terms of perplexity in
Table \ref{tab:valid_all} (second to last column, ``All Tasks MT'').
It achieves a {\em dodeca}Score of 19.1, superior to all pre-train fine-tune 
approaches except pushshift.io Reddit pre-training followed by fine-tuning, and is also superior
to a single pushshift.io Reddit model. However, comparing across tasks, while most are close 
to the corresponding best fine-tuned model, many are just slightly worse. This is an expected
result and is often reported in multi-task systems \citep{raffel2019exploring}.
We look upon this result as both positive -- we can obtain a single model doing well on all tasks, which a fine-tuned model cannot -- whilst also remaining  a challenge to
 the community: can one find architectures that leverage multi-tasking even better?

\paragraph{Multi-Task followed by Fine-Tuning} 
As also performed in \citet{liu2015representation,raffel2019exploring} 
we can try to train in a  multi-task manner on all
tasks, before fine-tuning on a single task, and build a separate model performing
this procedure for all tasks, in an attempt to improve single task results further.
Using this approach, one is free to perform hyperparameter search differently for each task.
Here, we found that applying relative task up-weighting during multi-tasking training
made a clear difference to the final quality of the fine-tuned target task model, see
Table \ref{tab:mt_ft_weighting_res}. Generally, better results come from assigning 
most of the multi-task weight towards the task itself to be fine-tuned.  Using such an approach 
we can get marginally better results than fine-tuning alone, although the differences
are generally small. 
The final best models per task are shown compared to other approaches in Table \ref{tab:valid_all} (third to last column, ``MT All Tasks + FT Single Task'').
The final validation {\em dodeca}Score is 16.8, only slightly below 17.1 for fine-tuning.

\paragraph{Decoding Strategies}
So far, we have only been measuring perplexity, but we are actually interested in generation, which requires us to decode. 
We consider several standard approaches:
greedy, beam search (with beam size, and minimum and maximum output length\footnote{The length parameters are important for ELI5.} hyperparameters), 
beam search with beam blocking (blocking $n$-grams, we use $n=3$) \citep{paulus2017deep} 
and nucleus sampling (with parameter $p$) \citep{holtzman2019curious}. 
We show the effect of these choices in Table \ref{tab:decoding_res} for ConvAI2 and Wizard of Wikipedia (WoW).

\
\paragraph{Final Systems} 
The final test performance for our best multi-task and fine-tuned (via multi-task followed by fine-tuning) systems are 
reported  in Table \ref{tab:sota} (right), with more
detailed results with all decoding-based metrics,
and validation as well as test performance
in 
Appendix \ref{app:moar_results}.
Here, for the multi-task model we have fine-tuned the decoding hyperparameters per task.
For results with a single set of decoding hyperparameters, see also Appendix \ref{app:moar_results}.
We generally find across all metrics a similar story as before when comparing the fine-tuning with multi-tasking: multi-tasking is successful, but the challenge is still to do better.

\paragraph{Comparison to Existing Systems} 

We compare to  existing state-of-the-art results previously  published for each task.
Results are given in Table \ref{tab:sota}. As existing works report different
metrics per task, we report perplexity where possible (but note, they may be computed on a different dictionary), and choose the sequence decoding-based metric that is commonly reported  per task (listed in column `Metric'), where the 'Score' column reports its value.
We compare these to our best fine-tuned and multi-tasked models.
Our multi-task model outperforms all available existing results, with 2 of the 12 tasks having no previous result. It is only surpassed by our fine-tuned model which also outperforms all available existing results. Overall, our methods set a strong challenge
to future approaches.

\paragraph{Human Evaluation}

In addition to automatic metrics, we perform human evaluation on two of the tasks to assess the abilities of our All Tasks MT conversational agent: the knowledge grounding task Wizard of Wikipedia (WoW) and the image grounding task Image Chat. We follow the same
evaluation protocols as in \citet{dinan2018wizard,shuster2018imagechat}, comparing our method to the existing approaches referenced in Table \ref{tab:sota}.
This involves collecting 100 human-bot conversations for WoW using crowdworkers, involving 8--10 turns each, across seen topics (seen in the training set) and unseen topics, 
and 500 image-based responses for Image Chat. A separate set of crowdworkers are then used
to compare models pairwise following the ACUTE-Eval procedure of \citep{li2019acute}, where they are asked to choose which is ``the more engaging response'' for Image Chat (1500 trials) and ``Who would you prefer to talk to for a long conversation?'' for WoW (400 trials).

\begin{figure}[t]
\begin{center}
\includegraphics[trim={0 0 0 3mm}, clip, width=0.48\textwidth]{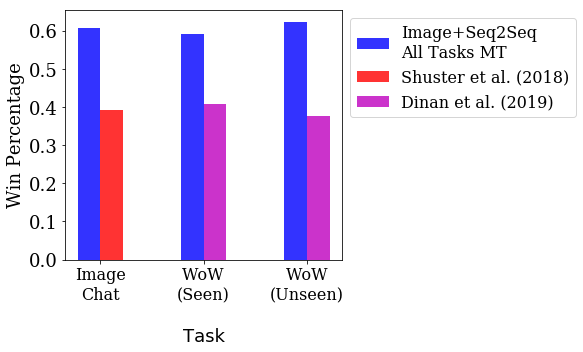}
\caption{Human evaluations on Image Chat and Wizard of Wikipedia (WoW), comparing existing state of the art models with our All Tasks MT  conversational agent. Engagingness win rates
 are statistically significant in all three matchups (binomial test, $p <.05$).}
\label{fig:human_eval}
\end{center}
\end{figure}

The results, given in Figure \ref{fig:human_eval}, show our method outperforming the existing
state of the art generative models 
on all three comparisons: Image Chat, WoW seen topics and WoW unseen topics.
All three results are statistically significant (binomial test, $p <.05$).
Additional details and results breakdown are given in Appendix Section \ref{app-sec:human_eval}.

\paragraph{Example Outputs} 

We show some example outputs of our multi-task model for some of the tasks in Appendix \ref{sec:model_examples}.
Our model is able to leverage images, knowledge, and given personality attributes
to produce engaging dialogue with a large amount of variety, depending on the situation.

\paragraph{Leave-One-Out Zero-Shot Performance} 
Last, but not least, we evaluate the performance of a multi-task model at zero-shot transfer
to a new dialogue task. This is performed by training on all but one of the tasks, and reporting performance on the left out one, repeating this experiment for all tasks.
Our best performing models in that regard are reported in Table \ref{tab:valid_all} (last column). First, it is reassuring that the overall scores are reasonable,
outperforming a pushshift.io Reddit only model on every task except pushshift.io Reddit itself. This means
that multi-tasking across many tasks helps transfer learning.
However, the gap between zero-shot performance and multi-task or fine-tuning performance means there
is still a significant challenge in improving these results.
Finally, we believe that reporting results in this regime  in addition to multi-tasking
results may help avoid the temptation to ``cheat'' at multi-tasking by trying to
detect the task and then apply a separate fine-tuned classifier, as presumably
 that approach will not truly leverage reasoning and skills between tasks, which
 transfer may help measure.

\section{Discussion}

We have introduced the {\em dodeca}Dialogue task, and provide strong baseline results
leveraging multi-modal Image+Seq2Seq transformers trained across all tasks.
The goal of introducing this task is not just as another challenge dataset, but 
to further motivate building and evaluating conversational agents capable of 
multiple skills -- one of the core goals of AI.
We believe current systems are closer to that goal than ever before -- 
but we also still have a long way to go. 

Recently reported results show systems  can be reasonably competitive
compared to humans 
in particular domains 
for short conversations \cite{li2019acute,shuster2018imagechat}. This work tries to bridge the gap to avoid agents
with niche skills, to move towards evaluating an open-domain set of skills.
Still, despite leveraging 12 tasks, there are many skills not included in our set.
For example, longer conversations involving memory \citep{moon-etal-2019-memory}, 
or mixing open-domain conversation
with task oriented goals. 
Future work should consider adding these tasks to the ones used here, while
continuing the quest for improved models.

\bibliography{acl2020}
\bibliographystyle{acl_natbib}

\newpage
\onecolumn
\appendix

\section{Additional Results}\label{app:moar_results}

 \begin{table*}[h!]
    \centering
    \begin{small}
    \setlength{\tabcolsep}{0.75em}
    \begin{tabular}{l|rrrrrr||rrrrrr}
   &  \multicolumn{6}{c}{MT + FT  } & \multicolumn{6}{c}{All Tasks MT} \\
    &  PPL  & BLEU & \multicolumn{3}{c}{ROUGE} & F1  &  PPL   & BLEU & \multicolumn{3}{c}{ROUGE} & F1\\
    &       &   4   & 1 & 2 & L &   &        &   4   & 1 & 2 & L & \\
ConvAI2 & 11.1 & 6.6 & 37.0 & 11.6 & 31.8 & 21.6 &  10.8 & 5.5 & 39.4 & 12.5 & 33.7 & 21.7 \\
DailyDialog & 10.4 & 4.0 & 35.6 & 10.0 & 30.8 & 18.2 &  12.0 & 2.9 & 33.9 & 8.7 & 29.2 & 16.2\\
Wiz. of Wikipedia  & 8.3 & 21.5 & 55.3 & 28.4 & 44.9 & 38.4 &  8.4 & 21.0 & 53.2 & 28.0 & 45.4 & 38.4 \\
Empathetic Dialog & 11.4 & 3.5 & 38.0 & 9.5 & 32.3 & 19.5 &  11.5 & 3.7 & 37.2 & 8.9 & 31.4 & 19.3 \\
Cornell Movie & 20.2 & 2.5 & 29.5 & 6.7 & 25.7 & 12.4 &  22.2 & 2.1 & 29.1 & 6.5 & 25.6 & 11.9 \\
LIGHT  & 18.9 & 2.6 & 30.8 & 5.8 & 24.8 & 16.2 &  19.3 & 2.4 & 30.5 & 5.6 & 24.6 & 16.1 \\
ELI5 &  21.0 & 3.7 & 38.6 & 7.2 & 22.1 & 23.1 &  24.9 & 3.2 & 35.2 & 6.3 & 20.5 & 21.3 \\
Ubuntu & 17.1 & 2.5 & 27.0 & 5.0 & 22.8 & 12.7 &  23.1 & 3.7 & 26.0 & 4.3 & 22.0 & 12.1 \\
Twitter & 30.7 & 3.2 & 16.5 & 3.3 & 14.3 & 9.9 &  38.2 & 2.6 & 19.4 & 3.3 & 16.5 & 9.8 \\
pushshift.io Reddit & 25.6 & 2.1 & 24.1 & 4.5 & 18.7 & 13.6 &  27.8 & 1.6 & 23.4 & 4.2 & 18.1 & 13.5 \\
Image Chat  & 18.8 & 2.4 & 30.1 & 5.7 & 26.0 & 13.0 &  22.3 & 2.1 & 28.4 & 4.9 & 24.6 & 12.9 \\
IGC & 11.9 & 8.6 & 65.0 & 34.1 & 60.5 & 38.4 &  12.0 & 8.0 & 61.3 & 28.3 & 56.8 & 41.4 \\
\hline
{\em dodeca}Score & 17.1 & 5.3 & 35.6 & 11.0 & 29.6 & 19.8 & 19.4 & 4.9 & 34.8 & 10.1 & 29.0 & 19.6 \\
    \end{tabular}
    \end{small}
    \caption{Test performance for various metrics on the {\em dodeca}Dialogue tasks comparing our multi-task and multi-task + fine-tuned methods. } 
    \label{tab:final_test_all}
    \centering
    \begin{small}
    \setlength{\tabcolsep}{0.75em}
    \begin{tabular}{l|rrrrrr||rrrrrr}
   &  \multicolumn{6}{c}{MT + FT  } & \multicolumn{6}{c}{All Tasks MT} \\
    &  PPL  & BLEU & \multicolumn{3}{c}{ROUGE} & F1  &  PPL   & BLEU & \multicolumn{3}{c}{ROUGE} & F1\\
    &       &   4   & 1 & 2 & L &   &        &    4  & 1 & 2 & L & \\
ConvAI2 &  11.2 & 5.7 & 36.7 & 10.9 & 31.6 & 21.1 &  11.3 & 5.3 & 38.7 & 11.6 & 32.9 & 21.3 \\
DailyDialog & 10.2 & 4.4 & 36.8 & 10.7 & 32 & 18.8 &  11.8 & 3.1 & 34.8 & 9.3 & 30.2 & 17.1 \\
Wiz. of Wikipedia  &  8.5 & 20.8 & 54.9 & 28.0 & 44.8 & 37.9 &  8.7 & 20.2 & 55.2 & 28.2 & 45.0 & 37.9 \\
Empathetic Dialog & 11.1 & 3.6 & 38.6 & 9.8 & 32.7 & 19.7 &  11.2 & 3.5 & 37.5 & 9.1 & 31.8 & 19.3 \\
Cornell Movie & 19.8 & 2.5 & 29.3 & 6.7 & 25.6 & 12.3 &  21.9 & 2.1 & 29.0 & 6.5 & 25.6 & 11.8 \\
LIGHT  & 18.7 & 2.6 & 31.2 & 6.2 & 25.2 & 16.5 &  19.0 & 2.5 & 30.9 & 6.1 & 25.0 & 16.4 \\
ELI5 & 21.1 & 3.7 & 38.7 & 7.3 & 22.1 & 23.2 &  25.0 & 3.2 & 35.3 & 6.3 & 20.6 & 21.2 \\
Ubuntu & 17.2 & 2.4 & 27.1 & 5.0 & 22.9 & 12.8 & 23.3 & 3.5 & 26.4 & 4.6 & 22.3 & 12.2 \\
Twitter & 29.8 & 3.2 & 16.7 & 3.5 & 14.5 & 10.1 &  37.0 & 2.6 & 19.7 & 3.6 & 16.8 & 9.9 \\
pushshift.io Reddit & 25.8 & 2.2 & 24.2 & 4.5 & 18.7 & 13.4 &  28.0 & 1.7 & 23.4 & 4.1 & 18.2 & 13.3 \\
Image Chat & 18.3 & 2.4 & 30.7 & 6.2 & 26.3 & 14.3 &  21.8 & 2.1 & 28.6 & 5.3 & 24.7 & 13.1 \\
IGC  & 10.0 & 10.6 & 67.9 & 38.2 & 64.5 & 45.1 &  10.2 & 11.0 & 66.3 & 34.8 & 61.4 & 45.3 \\
\hline
{\em dodeca}Score & 16.8 & 5.3 & 36.1 & 11.4 & 30.1 & 20.4 & 19.1 & 5.1 & 35.5 & 10.8 & 29.5 & 19.9 \\
    \end{tabular}
    \end{small}
    \caption{Validation performance for various metrics on the {\em dodeca}Dialogue tasks comparing our multi-task and multi-task + fine-tuned methods. } 
    \label{tab:final_valid_all}
\vspace{6mm}
    \centering
    \begin{small}
    \begin{tabular}{l|rrrrrr}
    &  PPL  & BLEU & \multicolumn{3}{c}{ROUGE} & f1\\
    &       &  4    &  1  &  2  &  L  &   \\
ConvAI2 & 11.3 & 5.6 & 22.2 & 7.0 &  20.4 & 21.3  \\
DailyDialog & 11.8 & 4.8 & 18.9 & 5.6 & 17.6 & 16.6 \\
Wiz. of Wikipedia  & 8.7 & 19.7 & 40.9 & 22.6 & 36.9 & 37.7 \\
Empathetic Dialog & 11.2 & 4.8 & 20.9 & 5.6 & 19.0 & 19.3 \\
Cornell Movie & 21.9 & 3.3 & 14.2 & 3.2 & 13.4 & 11.3 \\
LIGHT  & 19.0 & 2.9 & 17.0 & 3.4 & 15.0 & 16.2 \\
ELI5 & 25.0 & 1.6 & 14.2 & 2.6 & 9.6 & 16.2 \\
Ubuntu & 23.3 & 2.3 & 12.5 & 1.9 & 11.6 & 11.2 \\
Twitter & 37.0 & 2.3 & 9.5 & 1.7 & 8.7 & 8.9 \\
pushshift.io Reddit & 28.0 & 1.8 & 12.1 & 2.2 & 10.4 & 11.3 \\
Image Chat (all turns)  & 21.8 & 2.1 & 14.7 & 2.5 & 13.6 & 13.1\\
IGC  & 10.2 & 5.5 & 50.7 & 25.3 & 49.1 & 36.0 \\
\hline
{\em dodeca}Score & 19.1 & 4.7 & 20.7 & 7.0 & 18.8 & 18.3 \\
    \end{tabular}
    \end{small}
    \caption{All Tasks Multi-Tasking (MT) validation performance for various metrics on the {\em dodeca}Dialogue tasks with one set of decoding parameters: a beam size of 3, minimum response length of 10, and blocking repeated tri-grams.} 
    \label{tab:final_mt_onedecoding_valid_all}
\end{table*}
\newpage

\begin{table*}[t!]
\vspace{6mm}
    \centering
    \setlength{\tabcolsep}{0.2em}
    \begin{small}
    \begin{tabular}{l|rrrrc|rrrrc|rrrrc}
    &  \multicolumn{5}{c}{BLEU} & \multicolumn{5}{c}{ROUGE-L} & \multicolumn{5}{c}{F1}\\
    &  \tiny{Score} & \tiny{Beam} & \tiny{Min L} & \tiny{Max L} & \tiny{N-gram Block} &  \tiny{Score} & \tiny{Beam} & \tiny{Min L} & \tiny{Max L} & \tiny{N-gram Block}  & \tiny{Score} & \tiny{Beam} & \tiny{Min L} & \tiny{Max L} & \tiny{N-gram Block}\\
ConvAI2 & 5.7 & 10 & 10 & 128 & 3 & 31.6 & 10 & 50 & 128 & 3 & 21.1 & 3 & 10 & 128 & 3\\ 
DailyDialog & 4.4 & 10 & 5 & 128 & 3 & 32.0 & 3 & 50 & 128 & 3 & 18.8 & 5 & 10 & 128 & 3\\ 
Wiz. of Wikipedia  & 20.8 & 10 & 5 & 128 & 0 & 44.8 & 10 & 50 & 128 & 3 & 37.9 & 10 & 10 & 128 & 3 \\ 
Empathetic Dialog & 3.6 & 10 & 5 & 128 & 3 & 32.7 & 5 & 50 & 128 & 3 & 19.7 & 5 & 10 & 128 & 3\\ 
Cornell Movie & 2.5 & 10 & 5 & 128 & 3 & 25.6 & 10 & 50 & 128 & 3 & 12.3 & 10 & 20 & 128 & 3\\ 
LIGHT  & 2.6 & 3 & 5 & 128 & 3 & 25.2 & 5 & 50 & 128 & 3 & 16.5 & 5 & 20 & 128 & 3 \\ 
ELI5 & 3.7 & 10 & 200 & 256 & 3 & 22.1 & 5 & 200 & 256 & 3 & 23.2 & 10 & 200 & 256 & 3 \\ 
Ubuntu & 2.4 & 10 & 5 & 128 & 0 & 22.9 & 10 & 40 & 128 & 3 & 12.8 & 2 & 10 & 128 & 3 \\ 
Twitter & 3.2 & 10 & 20 & 128 & 3 & 14.5 & 5 & 50 & 128 & 3 & 10.1 & 10 & 20 & 128 & 3 \\ 
pushshift.io Reddit & 2.2 & 10 & 10 & 128 & 0 & 18.7 & 5 & 50 & 128 & 3 & 13.4 & 5 & 50 & 128 & 3 \\ 
Image Chat (all turns)  & 2.4 & 10 & 5 & 128 & 3 & 26.4 & 3 & 50 & 128 & 3 & 14.3 & 5 & 1 & 128 & 3 \\ 
IGC  & 10.6 & 10 & 5 & 128 & 3 & 64.5 & 3 & 50 & 128 & 3 & 45.1 & 10 & 5 & 128 & 3 \\ 
\hline
    \end{tabular}
    \end{small}
    \caption{Best decoding parameters for each task, based on metric. Scores are from the best performing task-specific multi-task + fine-tuned model on validation sets. "Min L" and "Max L" refer to the minimum and maximum decoding length, where "L" is the number of tokens.} 
    \label{tab:best_decode_params}
\end{table*}

\section{Human Evaluation Further Details}\label{app-sec:human_eval}
We provide additional results from our human evaluations described in Section \ref{sec:experiments}. In Figure \ref{fig:human_eval}, we compare our All Tasks MT Image+Seq2Seq model to existing baselines from both tasks; to produce those outputs, we used beam search with a beam size of 10 and tri-gram blocking. As with our experiments regarding automatic metrics, we additionally explored nucleus sampling, with parameter $p = 0.7$, and compared to both the baseline models as well as human outputs. In tables \ref{tab:image_chat_diagonal},  \ref{tab:wizard_seen_diagonal}, and \ref{tab:wizard_unseen_diagonal}, we show the full results of comparing various models both to each other and also to humans.

When collecting the model-human chats for Wizard of Wikipedia, we additionally asked the humans for a rating from 1-5 at the end of each conversation, to indicate the quality of the model's responses; we compare these Likert ratings to that of \citet{dinan2018wizard}, which followed the same protocol, in Table \ref{tab:wizard_likert}. The findings are similar to the pairwise ACUTE-Eval results in the main paper.

\begin{table*}[ht]
    \centering
    \setlength{\tabcolsep}{0.5em}
    \begin{small}
    \begin{tabular}{c|c|rrrr}
   & & \multicolumn{4}{c}{Win Percentage}\\
   \hline\hline
\multirow{5}{*}{Lose Percentage} & & \multicolumn{1}{c}{\cite{shuster2018imagechat}} & \multicolumn{1}{c}{Image+Seq2Seq} & \multicolumn{1}{c}{Image+Seq2Seq} & \multicolumn{1}{c}{Human} \\
& & & \multicolumn{1}{c}{Nucleus} & \multicolumn{1}{c}{Beam} & \\

& \cite{shuster2018imagechat} & - & 50.8 & $^*60.7$ & $^*79.3$ \\
& Image+Seq2Seq Nucleus & 49.2 & - & 52.1 & $^*73.8$ \\
& Image+Seq2Seq Beam & $^*39.3$ & 47.9 & - & $^*79.4$ \\
& Human & $^*20.7$ & $^*26.2$ & $^*20.6$ & -\\
\hline
    \end{tabular}
    \end{small}
    \caption{Human evaluations on Image Chat, comparing various decoding schemes for our Image+Seq2Seq model trained on all tasks MT, as well as comparisons with human outputs. Scores with $^*$ are statistically significant (binomial test, $p <.05$).} 
    \label{tab:image_chat_diagonal}
\end{table*}

\begin{table*}[ht]
    \centering
    \begin{small}
    \begin{tabular}{c|c|rrrr}
   & & \multicolumn{4}{c}{Win Percentage}\\
   \hline\hline
\multirow{5}{*}{Lose Percentage} & & \multicolumn{1}{c}{\cite{dinan2018wizard}} & \multicolumn{1}{c}{Image+Seq2Seq} & \multicolumn{1}{c}{Image+Seq2Seq} & \multicolumn{1}{c}{Human} \\
& & & \multicolumn{1}{c}{Nucleus} & \multicolumn{1}{c}{Beam} & \\

& \cite{dinan2018wizard} & - & 59.1 & 62.1 & 71.9 \\
& Image+Seq2Seq Nucleus & 40.1 & - & -  & 70.4 \\
& Image+Seq2Seq Beam & 37.9 & - & - & 60.0 \\
& Human & 28.1 & 29.6 & 40.0 & -\\
\hline
    \end{tabular}
    \end{small}
    \caption{Human evaluations on Wizard of Wikipedia (seen) test set, comparing various decoding schemes for our Image+Seq2Seq model trained on all tasks MT, as well as comparisons with human outputs, using ACUTE-Eval. All scores are statistically significant (binomial test, $p <.05$).} 
    \label{tab:wizard_seen_diagonal}
\end{table*}

\begin{table*}[t!]
    \centering
    \begin{small}
    \begin{tabular}{c|c|rrrr}
   & & \multicolumn{4}{c}{Win Percentage}\\
   \hline\hline
\multirow{5}{*}{Lose Percentage} & & \multicolumn{1}{c}{\cite{dinan2018wizard}} & \multicolumn{1}{c}{Image+Seq2Seq} & \multicolumn{1}{c}{Image+Seq2Seq} & \multicolumn{1}{c}{Human} \\
& & & \multicolumn{1}{c}{Nucleus} & \multicolumn{1}{c}{Beam} & \\

& \cite{dinan2018wizard} & - & 62.3 & 64.1 & 75.8 \\
& Image+Seq2Seq Nucleus & 37.7 & - & -  & 72.8 \\
& Image+Seq2Seq Beam & 35.9 & - & - & 60.5 \\
& Human & 24.2 & 27.2 & 39.5 & -\\
\hline
    \end{tabular}
    \end{small}
    \caption{Human evaluations on Wizard of Wikipedia (unseen) test set, comparing various decoding schemes for our Image+Seq2Seq model trained on all tasks MT, as well as comparisons with human outputs, using ACUTE-Eval. All scores are statistically significant (binomial test, $p <.05$).} 
    \label{tab:wizard_unseen_diagonal}
\begin{center}
\small
\begin{tabular}{llrr}
\toprule
Method & Decode Method & Seen &Unseen \\
\midrule
Human Performance            &       & 4.13 (1.08) & 4.34 (0.98)\\
\addlinespace[0.4em]
\cite{dinan2018wizard} & Beam  & 2.92 (1.33) & 2.93 (1.30) \\
\addlinespace[0.4em]
Image+Seq2Seq (All Tasks MT) & Beam & 3.42 (1.10) & 3.33 (1.09) \\
Image+Seq2Seq (All Tasks MT) & Nucleus &  2.95 (1.08) & 3.43 (1.05) \\
\bottomrule
\end{tabular}
\end{center}
\caption{Human evaluations on Wizard of Wikipedia, comparing various decoding schemes for our Image+Seq2Seq model trained on all tasks MT, as well as comparisons with human outputs, in terms of \textbf{Likert Scores}. Ratings are reported as mean (stddev).
}
\label{tab:wizard_likert}
\end{table*}

\newpage
\clearpage

\section{All Tasks Multi-Task Model Examples}\label{sec:model_examples}

\begin{figure*}[ht]
\small
\scriptsize
\begin{tabular}{p{5cm}rp{8cm}}

\toprule
Context & ConvAI2 \\
\midrule
\\\multirow{ 5 }{ 3cm }{ \textbf{{Personas:}}\\
 i love to drink wine and dance in the moonlight.\\
 i am very strong for my age.\\
 i'm 100 years old.\\
 i feel like i might live forever. }\\
& Speaker A: &

hi , how are you doing today ? \\
	& Speaker B: & good ! you ! celebrating with fellow centenarians \\
	& Speaker A: & nice . i'm playing some card games with my family . \\
	& Speaker B: & that sounds like fun . i like wine and dancing too ! \\
	& Speaker A: & same . i try to get a small workout in a three mile walk for me is key . \\
    \cmidrule(lr){2-2}
    \cmidrule(lr){2-2} \\
& Model: & that's great. i'm very strong for my age.\\ & & \\
 & & \\
\end{tabular}
\end{figure*}

\begin{figure*}[ht]
\small
\scriptsize
\begin{tabular}{p{5cm}rp{8cm}}

\toprule
Context & Wizard of Wikipedia \\
\midrule
\\\multirow{ 2 }{ 4cm }{ 
\textbf{Chosen Topic:}\\
The Rolling Stones\\
\textbf{Knowledge:}\\
 no\_passages\_used 
 }\\
	& Speaker A: & I love the Rolling Stones. Do you? They were formed in 1962. \\
	& Speaker B: & 
I love them!  Who was in the band at the start? \\
	& Speaker A: & The first lineup was Brian Jones, Mick, Keith, Bill Wyman, Charlie Watts and Ian Steward. \\
	& Speaker B: & 
Was Mick ever replaced? \\
    \cmidrule(lr){2-2}
    \cmidrule(lr){2-2} \\
& Model: & i'm not sure, but i do know that mick jagger was the lead vocalist.\\ & & \\
\end{tabular}
\end{figure*}

\begin{figure*}[ht]
\small
\scriptsize
\begin{tabular}{p{5cm}rp{8cm}}

\toprule
Context & Wizard of Wikipedia \\
\midrule
\\\multirow{ 2 }{ 4cm }{ 
\textbf{Chosen Topic:}\\
Dog\\
\textbf{Knowledge:}\\
 The dog was the first species to be domesticated and has been selectively bred over millennia for various behaviors, sensory capabilities, and physical attributes. 
 }\\
& Speaker A: & 

Hi, do you like dogs? \\
	& Speaker B: & I do! Actually, my favourite is the Basenji breed! It is a hunting dog. \\
	& Speaker A: & 
Oh, those are cool. I have seen them on the Westminster Dog Show, but I've never met one in person. I have a Havanese.  \\
	& Speaker B: & Wow, are you Cuban? I know that the Havanese are the national dog of cuba! \\
	& Speaker A: & 
No, not Cuban, but yes they are the national dog of Cuba! Apparently all the Havanese in the US are descended from a handful that came here from Cuba during the revolution! \\
    \cmidrule(lr){2-2}
    \cmidrule(lr){2-2} \\
& Model: & that's so cool! the dog was the first species to be domesticated!\\ & & \\
\end{tabular}
\small
\scriptsize
\begin{tabular}{p{5cm}rp{8cm}}

\toprule
Context & Wizard of Wikipedia \\
\midrule
\\\multirow{ 2 }{ 4cm }{
\textbf{Chosen Topic:}\\
Orc\\
\textbf{Knowledge:}\\
 While the overall concept of orcs draws on a variety of pre-existing mythology, the main conception of the creatures stems from the fantasy writings of J. R. R. Tolkien, in particular "The Lord of the Rings".
 }\\
 \\
	& Speaker A: & Orcs are cool fictional humanoid beings \\
	& Speaker B: &
Yeah I've seen them in a lot of things like Magic and DnD. \\
	& Speaker A: & While the concept remains the same it draws on many preexisting mythology  \\
	& Speaker B: &
That's really interesting! Orcs are some of my favorite fictional characters honestly. \\
    \cmidrule(lr){2-2}
    \cmidrule(lr){2-2} \\
& Model: & the main conception of the creatures stems from the fantasy writings of the lord of the rings\\ & & \\
\\

\end{tabular}
\end{figure*}

\begin{figure*}[ht]
\small
\scriptsize
\begin{tabular}{p{5cm}rp{8cm}}

\toprule
Context & Empathetic Dialogues \\
\midrule
\\\multirow{ 6 }{ 4cm }{
\textbf{Situation}\\
I received concert tickets for Christmas.\\
\textbf{Emotion}\\
excited
 }\\
& Speaker A: & I couldn't wait to go to the concert. \\
	& Speaker B: & Which Concert? \\
	& Speaker A: & The U2 concert. Tickets were really expensive and I never thought we would be able to go, but somehow we did!!! \\
    \cmidrule(lr){2-2}
    \cmidrule(lr){2-2} \\
& Model: & that's awesome. i've never been to a u2 concert.\\ & & \\
 & & \\
 & & \\
\\

\end{tabular}
\end{figure*}

\begin{figure*}[ht]
\small
\scriptsize
\begin{tabular}{p{5cm}rp{8cm}}

\toprule
Context & Empathetic Dialogues \\
\midrule
\\\multirow{ 6 }{ 4cm }{
\textbf{Situation}\\
I went to buy a 42 inch tv at walmart, the billboard says 299 usd, it was at 399 usd after taxes.\\
\textbf{Emotion}\\
disappointed
 }\\
& Speaker A: & So yeah i drove 2 hours to get to wallmart. with 299 in my wallet as advertised, and the real price of the tv was 399 usd. what a bummer \\
	& Speaker B: & Thats horrible. \\
	& Speaker A: & Yeah i had my hopes reall high and then i left empty handed, such a terrible feeling! \\
    \cmidrule(lr){2-2}
    \cmidrule(lr){2-2} \\
& Model: & i know what you mean. i would have been so mad.\\ & & \\
 & & \\

\end{tabular}
\end{figure*}

\begin{figure*}[ht]
\small
\scriptsize
\begin{tabular}{p{5cm}rp{8cm}}

\toprule
Context & Empathetic Dialogues \\
\midrule
\\\multirow{ 6 }{ 4cm }{
\textbf{Situation}\\
All my friends live in a different country\\
\textbf{Emotion}\\
lonely
 }\\
& Speaker A: & hi, i feel so lonely sometimes because all my friends live in a different country \\
	& Speaker B: & Oh, I'm sure you are lonely. Maybe you can join some kind of club that lets you meet new friends? \\
	& Speaker A: & i was thinking about it! I wanted to join a group for local moms \\
    \cmidrule(lr){2-2}
    \cmidrule(lr){2-2} \\
& Model: & that's a great idea! i'm sure you'll meet some great people there.\\ & & \\
 & & \\

\end{tabular}
\end{figure*}

\begin{figure*}[ht]
\small
\scriptsize
\begin{tabular}{p{5cm}rp{8cm}}

\toprule
Context & Cornell Movie \\
\midrule
\\\multirow{1}{ 5cm }{No Context}
& Speaker A: & Why'd you let him get to you? \\
	& Speaker B: & Who? \\
	& Speaker A: & Dorsey. \\
	& Speaker B: & I hate him. \\
	& Speaker A: & I know.  It'd have to be a pretty big deal to get you to mainline tequila. You don't seem like the type. \\
    \cmidrule(lr){2-2}
    \cmidrule(lr){2-2} \\
& Model: & i'm not above all that. i just don't think i can be "laid back" like everyone else.\\ & & \\
\\

\end{tabular}
\end{figure*}

\begin{figure*}[ht]
\small
\scriptsize
\begin{tabular}{p{5cm}rp{8cm}}

\toprule

Context & LIGHT \\
\midrule
\\\multirow{ 15 }{ 5cm }{ 
 \textbf{ \_setting\_name }: Castle Maids' Room, Inside Castle\\
 \textbf{ \_setting\_desc }: Inside the maids room is dark and grey.  It is much different from the rest of the Castle which has grand walls and floors and paintings from famous painters around the wall.  The maid holds many secrets in this room.  One which is a secret door that holds paintings she should not have.\\
 \textbf{ \_partner\_name }: painter\\
 \textbf{ \_self\_name }: worker\\
 \textbf{ \_self\_persona }: I am a worker from a village up the river. I tie the logs our woodsmen fell into rafts, and steer them down to the next town for milling. I am hardworking and patient. I enjoy fishing off of the edge of the raft during quiet times on the raft.\\
  }\\
& Speaker A: & Hello \\
& Speaker B: & Good Evening. You must be the painter of this room. \\
	& Speaker A: &  Yes sir! do you have a preferred color?
\_self\_act get brush
 \\
	& Speaker B: & I am just a lowly worker. I think the red paint looks good. I can assist you with painting. \\
	& Speaker A: &  I will love that. Tell me, how long have you been here?
\_partner\_act give brush to worker
\_self\_act drop paint brushes
 \\
    \cmidrule(lr){2-2}
    \cmidrule(lr){2-2} \\
& Model: & i've been here for a long time. i have been working here since i was a child.\\ & & \\
 & & \\
 & & \\
 & & \\
\\

\end{tabular}
\end{figure*}

\begin{figure*}[ht]
\small
\scriptsize
\begin{tabular}{p{5cm}rp{8cm}}

\toprule
Context & Daily Dialog \\
\midrule
\\\multirow{ 2 }{ 2cm }{
\textbf{Topic}\\
ordinary\_life\\
\textbf{Emotion}\\
no\_emotion
 }\\
& Speaker A: & So how did I do on my driving test ? \\
	& Speaker B: & Do you want the truth ? \\
	& Speaker A: & Of course , I do . \\
	& Speaker B: & Well , you really didn't do all that well . \\
	& Speaker A: & How bad did I do ? \\
    \cmidrule(lr){2-2}
    \cmidrule(lr){2-2} \\
& Model: & don't worry, i'll give you another try.\\ & & \\
\end{tabular}
\end{figure*}

\begin{figure*}[ht]
\small
\scriptsize
\begin{tabular}{p{5cm}rp{8cm}}

\toprule
Context & pushshift.io  Reddit \\
\midrule
\\\multirow{ 2 }{ 2cm }{
\textbf{Subreddit}\\
beer\\
 }\\
& Speaker A: & In Santa Cruz? Be SURE to hit up Sante Adairius Rustic Ales in Capitola! If you like farmhouse, funky, sour ales, they are real tough to beat. Their IPAs and porters are also amazing. Another great new brewery is Discretion, in Capitola, and I am also a fan of Santa Cruz Mountain Brewing and Seabright Brewing for affordability's sake (\$5 pitchers at Seabright on Tuesday/Thursday I believe?) but some beer snobs would turn their noses up at those places. Also, burger in Santa Cruz and Aptos always has a great taplist, as well as Aptos BBQ in Aptos. \\
    \cmidrule(lr){2-2}
    \cmidrule(lr){2-2} \\
& Model: & thanks for the suggestions! i'll have to check out sante adairius rustic ales. i've heard good things about seabright brewing, but haven't had a chance to try it yet. do you know much about it?\\ 
\end{tabular}
\end{figure*}

\begin{figure*}[ht]
\small
\scriptsize
\begin{tabular}{p{5cm}rp{8cm}}

\toprule
Context & Twitter \\
\midrule
\\\multirow{1}{ 5cm }{No Context}
& Speaker A: & reading now . this is absolutely fantastic . \\
    \cmidrule(lr){2-2}
    \cmidrule(lr){2-2} \\
& Model: & thank you so much. i'm glad you like it. i hope you enjoy it as much as i do.\\ & & \\
\end{tabular}
\end{figure*}

\begin{figure*}[ht]
\small
\scriptsize
\begin{tabular}{p{5cm}rp{8cm}}

\toprule
Context & Ubuntu \\
\midrule
\\\multirow{1}{ 5cm }{No Context}
& Speaker A: & i set up my hd such that i have to type a passphrase to access it at boot. how can i remove that passwrd, and just boot up normal. i did this at install, it works fine, just tired of having reboots where i need to be at terminal to type passwd in. help? . \\
& Speaker B: & backup your data, and re-install without encryption "might" be the easiest method . \\
    \cmidrule(lr){2-2}
    \cmidrule(lr){2-2} \\
& Model: & i'll try that, but i'm not sure how to do it without encryption..\\ & & \\
\end{tabular}
\end{figure*}

\begin{figure*}[ht]
\small
\scriptsize
\begin{tabular}{p{5cm}rp{8cm}}

\toprule
Context & ELI5 \\
\midrule
\\\multirow{ 10 }{ 5cm }{ 
\textbf{Knowledge}\\
$<$P$>$ And it does n’t come without questioning and critical thought . What we feel entitled to is a better world . And the thing is , we can see it , clearer and with broader scope than anyone before us , because all we know is the entire world , in real time . $<$P$>$ Page 1 of 2 1 2 Next $>$ Spycke83 PSLS Level : Bronze Messages : 52 Likes Received : 20 PSN : Spycke83 I`m wondering what you guys think of this newish trend of gamers feeling they are entitled certain things from a game . . .\\
 }\\
& Speaker A: & Why do recent generations feel so entitled , like everyone owes them ? --T--  \\
    \cmidrule(lr){2-2}
    \cmidrule(lr){2-2} \\
& Model: & it's not that they feel entitled, but that they don't have enough money to pay for it.\\ & & \\
 & & \\
 & & \\
 & & \\
 & & \\
 & & \\
 & & \\
 & & \\
 & & \\
 & & \\
 & & \\
 & & \\
 & & \\
 & & \\
 & & \\
 & & \\
 & & \\
 & & \\
 & & \\
 & & \\
\\

\end{tabular}
\end{figure*}

\begin{figure*}[ht]
\small
\scriptsize
\begin{tabular}{p{5cm}rp{8cm}}

\toprule
Context & Image Chat \\
\midrule
\\\multirow{ 6 }{ 2cm }{ 
\textbf{Image}\\
\includegraphics[height=25ex, width=38ex]{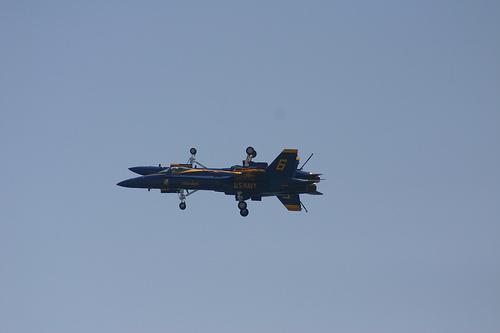}\\
\textbf{Personality}\\
Emotional
 }\\
& Speaker A: & Airplanes are scary to get on, you never know if it will crash or not. \\
	& Speaker B: & But these are professional pilots though. \\
    \cmidrule(lr){2-2}
    \cmidrule(lr){2-2} \\
& Model: & i can't imagine being a professional pilot.\\ & & \\
 & & \\
 & & \\
 & & \\
 & & \\
 & & \\
 & & \\
 & & \\

\end{tabular}
\end{figure*}

\begin{figure*}[ht]
\small
\scriptsize
\begin{tabular}{p{5cm}rp{8cm}}

\toprule
Context & Image Chat \\
\midrule
\\\multirow{ 6 }{ 2cm }{ 
\textbf{Image}\\
\includegraphics[height=25ex, width=38ex]{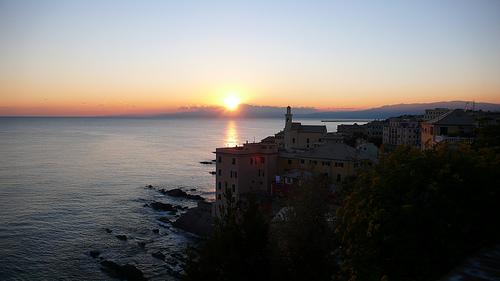}\\
\textbf{Personality}\\
Playful
 }\\
& Speaker A: & Great for night fishing. \\
	& Speaker B: & Meh, I doubt any fish would bite anyway. \\
    \cmidrule(lr){2-2}
    \cmidrule(lr){2-2} \\
& Model: & i m sure there are plenty of fish that would bite.\\ & & \\
 & & \\
 & & \\
 & & \\
 & & \\
 & & \\
 & & \\
\\
\end{tabular}
\end{figure*}

\begin{figure*}[ht]
\small
\scriptsize
\begin{tabular}{p{5cm}rp{8cm}}

\toprule
Context & Image Chat (First Turn) \\
\midrule
\\\multirow{ 6 }{ 2cm }{ 
\textbf{Image}\\
\includegraphics[height=25ex, width=38ex]{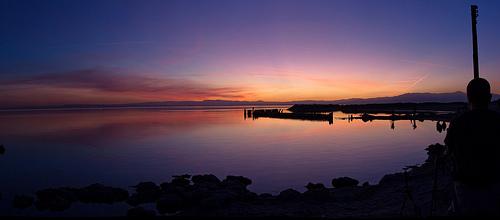}\\
\textbf{Personality}\\
Curious
 }\\
& Model: & i wonder if this is a lake?\\ & & \\
 & & \\
 & & \\
 & & \\
 & & \\
 & & \\
 & & \\
 & & \\
 & & \\
\\
\end{tabular}
\end{figure*}

\begin{figure*}[ht]
\small
\scriptsize
\begin{tabular}{p{5cm}rp{8cm}}

\toprule
Context & IGC \\
\midrule
\\\multirow{ 6 }{ 2cm }{
\textbf{Image}\\
\includegraphics[height=25ex, width=38ex]{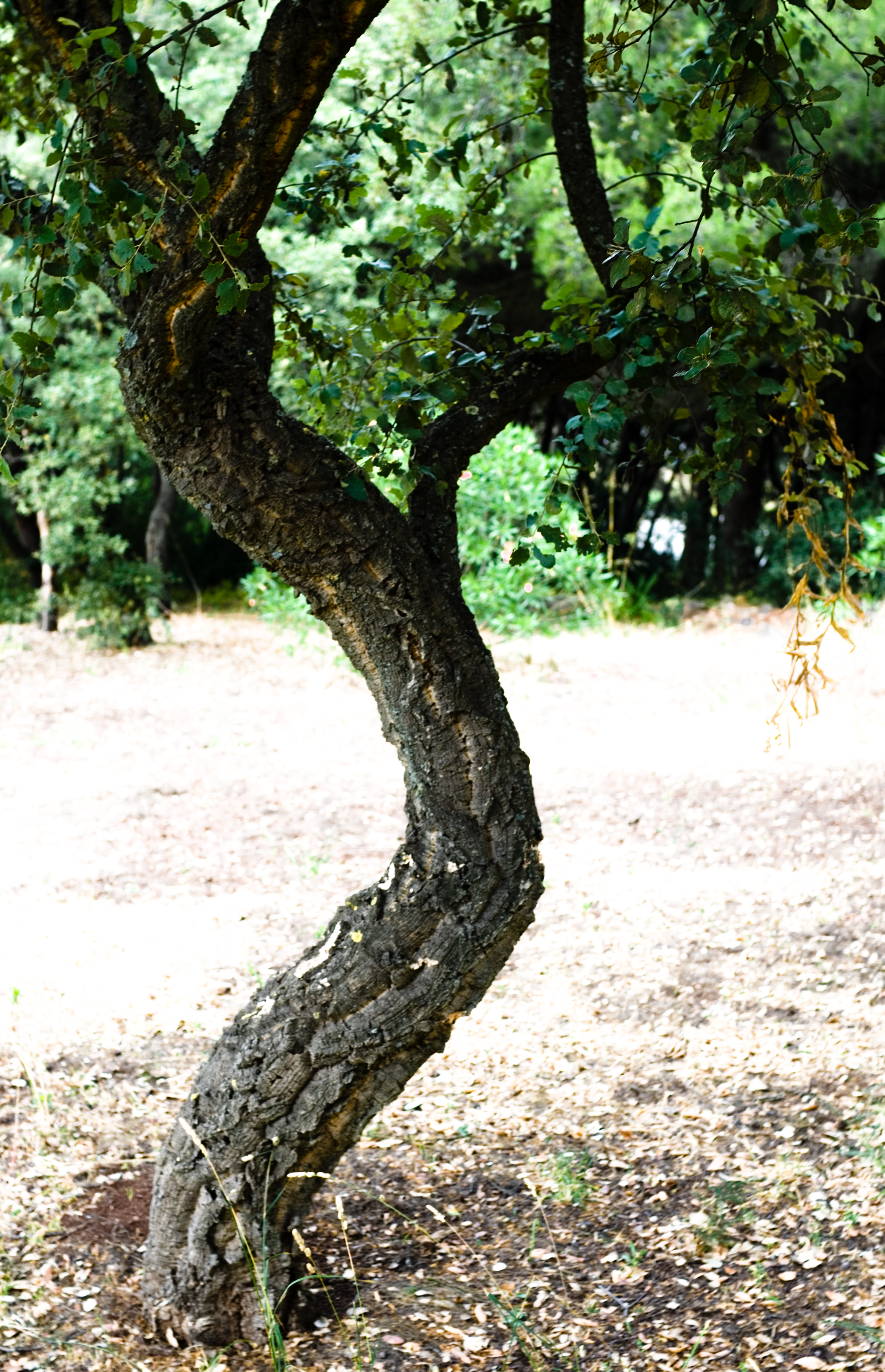}\\
 }\\
& Speaker A: & I visited this forest on a walk. \\
	& Speaker B: & Was this tree special? \\
    \cmidrule(lr){2-2}
    \cmidrule(lr){2-2} \\
& Model: & yes, it's one of the oldest trees in the world.\\ & & \\
 & & \\
 & & \\
 & & \\
 & & \\
 & & \\
\\
\end{tabular}
\end{figure*}

\begin{figure*}[ht]
\small
\scriptsize
\begin{tabular}{p{5cm}rp{8cm}}

\toprule
Context & IGC \\
\midrule
\\\multirow{ 6 }{ 2cm }{
\textbf{Image}\\
\includegraphics[height=25ex, width=38ex]{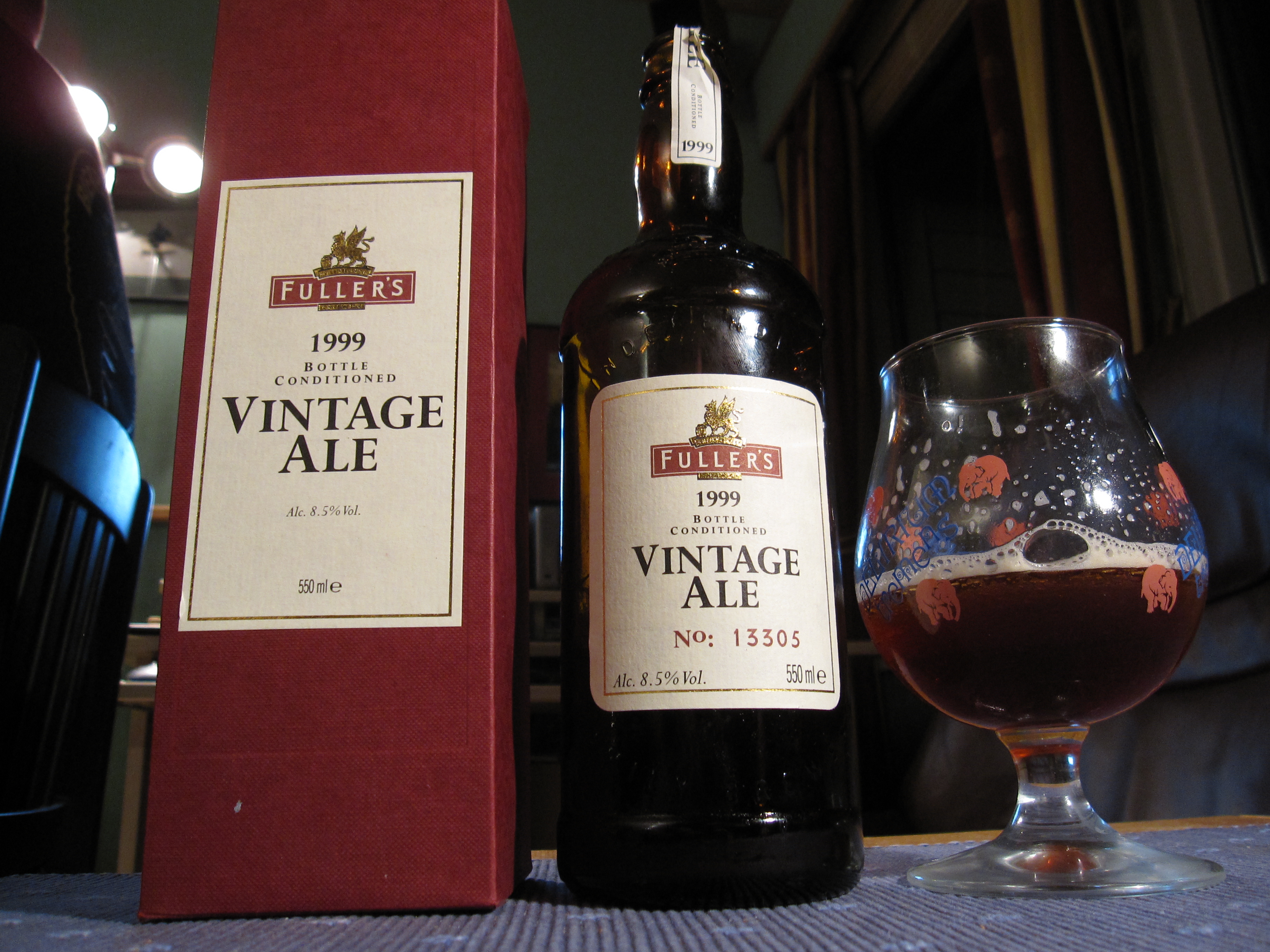}\\
 }\\
& Speaker A: & This is what I am sipping on today. \\
	& Speaker B: & I'm jealous, how much did it cost? \\
    \cmidrule(lr){2-2}
    \cmidrule(lr){2-2} \\
& Model: & it cost me \$20.\\ & & \\
 & & \\
 & & \\
 & & \\
 & & \\
 & & \\
\\
\bottomrule
\end{tabular}
\end{figure*}

\end{document}